\documentclass[conference]{IEEEtran}

\usepackage{cite}
\usepackage{amsmath,amssymb,amsfonts}
\usepackage{algorithmic}
\usepackage{graphicx}
\usepackage{textcomp}
\usepackage{placeins}
\usepackage{xcolor}
\usepackage{booktabs}
\usepackage{multirow}
\usepackage{tabularx}
\usepackage{url}
\usepackage{stfloats}
\usepackage{tikz}
\usetikzlibrary{positioning,arrows.meta,shapes.geometric}
\def\BibTeX{{\rm B\kern-.05em{\sc i\kern-.025em b}\kern-.08em
    T\kern-.1667em\lower.7ex\hbox{E}\kern-.125emX}}

\begin{document}


\title{Evaluating Multi-Task Morphological Concept Learning for Pulmonary Nodule Malignancy Assessment in 3D CT}


\author{
\IEEEauthorblockN{Namitha Narayanan}
\IEEEauthorblockA{
\textit{Independent Researcher} \\
Kerala, India \\
Email: namin232@gmail.com
}
}

\maketitle


\begin{abstract}
Morphological characteristics such as spiculation and lobulation play an important role in assessing pulmonary nodules on computed tomography (CT), particularly in relation to malignancy risk. This study examines whether learning radiologist-annotated morphological features together with malignancy risk from lesion-centred 3D CT volumes improves classification performance. The Lung Image Database Consortium and Image Database Resource Initiative (LIDC-IDRI) dataset was used, comprising 3,918 reader-level nodule annotations from 742 patients after excluding indeterminate malignancy ratings. Patient-level splitting was used for training, validation, and testing, with 112 patients and 628 reader annotations in the held-out test set. A single-task 3D convolutional neural network was compared with a multi-task model predicting malignancy risk, spiculation, and lobulation. The single-task model achieved a balanced accuracy of 0.548 and receiver operating
characteristic area under the curve (ROC-AUC) of 0.552, while the multi-task model achieved 0.539 and 0.558, respectively. Patient-level bootstrap analysis showed an ROC-AUC difference of 0.005 (95\% confidence interval (CI): -0.087 to 0.090) and a balanced-accuracy difference of -0.009 (95\% CI: -0.067 to 0.043). The auxiliary tasks were strongly imbalanced and showed limited predictive performance. Overall, including morphological features did not clearly improve malignancy-risk classification, showing the importance of class balance, label formulation, and reader-level annotation structure in multi-task pulmonary CT analysis.
\end{abstract}
 



\section{Introduction}

Pulmonary nodule assessment is an important part of identifying lesions that
may carry an increased risk of malignancy. Computed tomography (CT) provides
volumetric imaging of the thorax and allows pulmonary nodules to be examined
across multiple slices, making it possible to assess their size, shape,
margins, and other radiological characteristics in greater detail. Low-dose
CT has also become an established approach for lung-cancer screening in
high-risk populations, supported by evidence showing a reduction in
lung-cancer mortality through CT-based screening \cite{lancaster2022ldct}.
However, imaging-based assessment contributes to malignancy-risk estimation
rather than providing histopathological confirmation of disease.

Estimating malignancy risk from CT remains challenging because the labels used
for training and evaluation may themselves contain uncertainty. In the Lung Image Database Consortium and Image Database Resource Initiative
(LIDC-IDRI),
malignancy likelihood and morphological characteristics such as spiculation
and lobulation are assigned by radiologists rather than uniformly confirmed
through histopathology \cite{dai2023msnet}. Pulmonary nodules can also vary
considerably in their radiological appearance, and different readers may
interpret the same characteristics differently. Previous analysis of
LIDC-IDRI annotations has reported substantial interobserver disagreement in
several diagnostic characteristics, including malignancy, spiculation, and
lobulation \cite{lin2017interobserver}. These annotations therefore provide
valuable information about radiological assessment, but they should not be
treated as uncertainty-free ground truth.

Morphological characteristics can provide useful information when assessing
whether a pulmonary nodule appears more or less suspicious. Features such as
spiculation and lobulation describe aspects of the lesion border and shape
and have previously been associated with malignancy-risk assessment
\cite{chen2022guideline,liu2017traits}. In the present study, these two
morphological characteristics were therefore selected as auxiliary targets
in the multi-task model to examine whether learning them alongside malignancy
risk could provide additional predictive information and context for the
primary classification task.

Deep-learning models are well suited to medical image classification because
they can learn complex image representations directly from volumetric data.
For pulmonary nodule assessment, three-dimensional convolutional neural
networks (3D CNNs) can use information across multiple CT slices rather than relying on
a single two-dimensional view. In addition to predicting malignancy risk as a
single output, related imaging characteristics can also be introduced as
auxiliary prediction tasks. Previous work has used nodule density and
morphological characteristics as auxiliary labels in multi-task models for
pulmonary nodule classification \cite{wang2022deepln}. In the present study,
the single-task and multi-task models use the same 3D encoder, while the
multi-task model adds separate prediction heads for spiculation and lobulation.
This allows the experiment to examine whether the addition of these
morphological tasks provides any benefit to malignancy-risk prediction when
the underlying encoder is kept unchanged.

Previous studies have explored the use of semantic or morphological
characteristics alongside malignancy prediction in deep-learning models.
This is relevant because radiologists assess pulmonary nodules not only by an
overall malignancy score, but also through individual imaging characteristics
such as shape and margin appearance. Incorporating such characteristics into
a model may therefore provide additional context alongside the primary
malignancy-risk prediction. Existing approaches have used semantic prediction
together with additional architectural or feature-fusion components
\cite{shen2019semantic}. In the present study, a simpler comparison is used:
the single-task and multi-task models share the same 3D encoder, while the
multi-task model adds separate prediction heads for spiculation and
lobulation. This allows the effect of the auxiliary morphological tasks to be
examined more directly. At the same time, reader-level variability and class
imbalance in these annotations may limit how effectively the auxiliary tasks
can be learned.

The present study implements both a single-task and a multi-task model to
examine whether learning morphological characteristics alongside
malignancy risk provides a measurable benefit for 3D pulmonary nodule
classification from CT data. To ensure that the comparison primarily reflects
the inclusion or exclusion of the auxiliary morphological tasks, the same
underlying 3D CNN encoder is used in both models. The multi-task model
additionally predicts spiculation and lobulation through separate auxiliary
heads, while malignancy-risk prediction remains the primary task. The study
also assesses uncertainty in the model comparison using patient-level
bootstrap resampling and explores whether model performance varies across CT
slice-thickness groups.


\section{Related Work}

Deep-learning approaches have been widely investigated for the detection and
classification of pulmonary nodules from CT images, with three-dimensional
convolutional neural networks providing a way to learn volumetric information
across adjacent image slices \cite{halder2021survey,siddiqui2023_3dcnn}.
Recent studies have continued to evaluate 3D architectures for distinguishing
benign and malignant pulmonary nodules using volumetric CT data. Building on
this established use of 3D CNNs, the present study uses a common 3D encoder to
evaluate whether adding morphological concept learning through auxiliary
prediction tasks can provide additional benefit for pulmonary nodule
malignancy-risk classification.

Semantic and morphological characteristics have also been incorporated into
deep-learning models for pulmonary nodule malignancy assessment. Shen et al.
proposed a hierarchical semantic CNN in which radiologist-interpreted nodule
characteristics were learned as intermediate outputs and then incorporated
into the final malignancy prediction \cite{shen2019semantic}. More recently,
Wang et al. used nodule density and morphological characteristics as auxiliary
tasks in a multi-task 3D model for benign--malignant classification
\cite{wang2022deepln}. These studies support the idea that semantic information
can be learned alongside malignancy prediction within a deep-learning
framework. This provides a relevant basis for examining selected
morphological characteristics as auxiliary targets in the present study,
while also retaining outputs that are more closely related to features used
in radiological assessment.

The present study extends previous work on pulmonary nodule malignancy-risk
prediction by using a simple and controlled comparison between a single-task
model and a multi-task model that share the same 3D CNN encoder. The main
difference between the two models is that the multi-task model adds separate
prediction heads for spiculation and lobulation, which were selected as the
morphological characteristics of interest in this experiment. This design
allows the study to examine more directly whether learning these selected
morphological features provides additional predictive value for malignancy-risk
classification without changing the underlying feature extractor. In addition,
the study considers patient-level uncertainty and CT slice-thickness variation,
providing further context for interpreting model performance with respect to
reader-level annotation structure and acquisition heterogeneity.


\section{Materials and Methods}


\subsection{Dataset and Cohort Definition}

This study was performed using the Lung Image Database Consortium and Image Database Resource Initiative (LIDC-IDRI) dataset \cite{lidc_dataset,armato2011lidc}. The dataset is publicly available through The Cancer Imaging Archive (TCIA) \cite{lidc_dataset}. These annotations include morphological characteristics as well as malignancy ratings ranging from 1 to 5. After parsing the reader annotations, the corresponding malignancy rating was assigned to each reader-level nodule record. Ratings of 1 and 2 were considered lower malignancy risk, while ratings of 4 and 5 were considered higher malignancy risk. To reduce ambiguity in the binary classification experiment, rating 3 was excluded. This exclusion should not be interpreted as improving clinical reliability, since indeterminate cases remain an important consideration in real clinical settings.

The resulting cohort contained 3,918 reader-level nodule annotations obtained from 742 patients. Separate training, validation, and test sets were created using patient-level splitting. The training set contained 519 patients and 2,754 annotations, the validation set contained 111 patients and 536 annotations, and the held-out test set contained 112 patients and 628 annotations. A fixed random seed of 42 was used to support reproducibility of the patient assignment.

It is important to note that the experimental unit was a reader-level nodule annotation rather than a consensus physical lesion. Different radiologists may therefore provide separate annotations for the same pulmonary nodule \cite{armato2011lidc}. For this reason, patient-level separation was maintained across the training, validation, and test sets to prevent patient overlap and data leakage between the three dataset partitions.


\subsection{Label Definition}

The final classes associated with the primary single-task prediction were low malignancy risk and high malignancy risk, based on radiologist-assessed lesion annotations provided in the Extensible Markup Language (XML) files. It is important to note that the study estimates the risk of a lesion being malignant purely from reader-level nodule information provided by the radiologists, and the labels were not based on histopathological confirmation \cite{armato2011lidc}.

Lesions with malignancy ratings of 1 or 2 were assigned to the negative class and considered lower risk, whereas ratings of 4 or 5 were assigned to the positive class and considered higher risk \cite{armato2011lidc}. Rating 3 was excluded because it represented an indeterminate category and the experiment was formulated as a binary classification task.

For the multi-task model, two morphological features, spiculation and lobulation, were predicted alongside malignancy risk. The absence or indeterminate status of an auxiliary morphological label did not determine whether a lesion could still be used for malignancy-risk prediction. If the spiculation or lobulation rating was 3, or if the corresponding information was unavailable, that auxiliary label was masked and excluded from the respective auxiliary loss. The malignancy-risk prediction was still retained for that sample. The same low-risk and high-risk malignancy label definition used in the single-task model was also applied to the multi-task model.


\subsection{CT Preprocessing and Lesion Extraction}
A three-dimensional CT volume was reconstructed from the Digital Imaging and
Communications in Medicine (DICOM) image series. The DICOM slices were ordered using the patient position and orientation metadata, primarily \textit{ImagePositionPatient} and \textit{ImageOrientationPatient}, so that the individual slices could be stacked in the correct anatomical order \cite{dicomstandard}. The stored DICOM pixel values were then converted to Hounsfield units (HU) by multiplying each pixel value by the corresponding rescale slope and adding the rescale intercept values \cite{dicomstandard}.

The XML annotations were subsequently parsed to obtain the radiologist-defined regions of interest (ROIs). Each XML contour contained two-dimensional coordinates together with the image
Service-Object Pair Unique Identifier (SOP UID) identifying the corresponding CT slice \cite{armato2011lidc}. This SOP UID was matched against the DICOM \textit{SOPInstanceUID} so that the reader-provided contours could be aligned with the reconstructed CT volume. When an exact SOP UID match was not available, geometric slice position was used as a fallback. Through this alignment, the two-dimensional radiologist contours were mapped to their corresponding three-dimensional lesion location.

The centre of each lesion was estimated from the mean position of the included ROI contour points. A spacing-aware three-dimensional crop was then extracted around this centre. The extracted volume was resampled to an isotropic voxel spacing of $1 \times 1 \times 1$ mm using trilinear interpolation, producing a fixed input size of $64 \times 64 \times 64$ voxels. Regions extending outside the available CT volume were padded using $-1000$ HU.

Finally, the Hounsfield values were clipped to the range $[-1000, 400]$ HU and min--max normalized to $[0,1]$, consistent with preprocessing approaches used in pulmonary nodule deep-learning studies \cite{xiao2019mshcnn}.

\begin{figure}[htbp]
\centering
\includegraphics[width=\columnwidth]{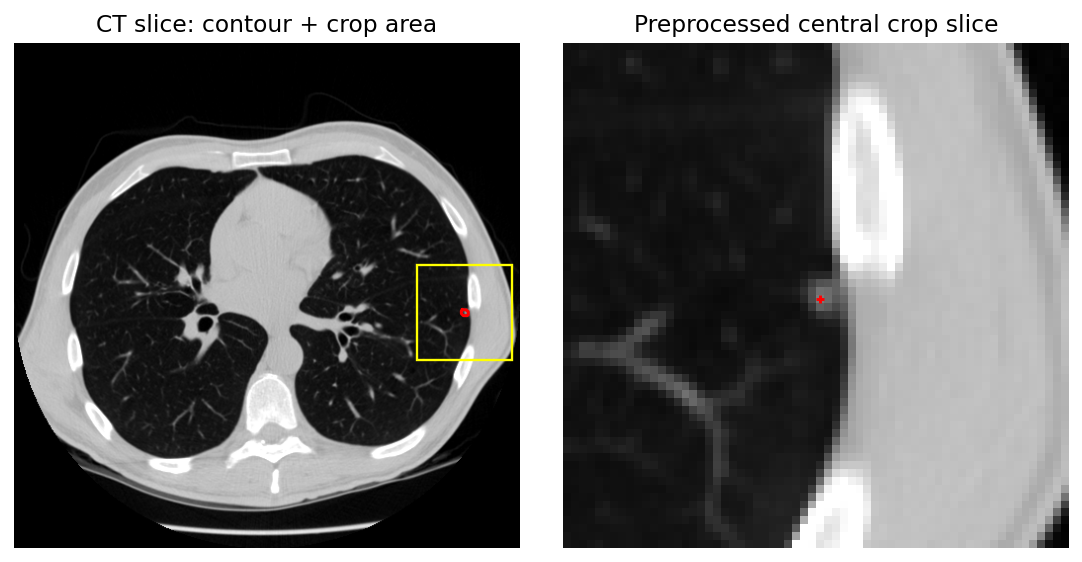}
\caption{Example of lesion localisation and crop extraction from LIDC-IDRI CT data. The left panel shows the original CT slice with the radiologist-defined lesion contour and the corresponding crop region. The right panel shows the central slice of the preprocessed lesion-centred crop used as model input.}
\label{fig:lesion_crop}
\end{figure}

An example of the lesion localisation and crop-extraction procedure is shown in Fig.~\ref{fig:lesion_crop}.

\subsection{Single-Task 3D CNN}

For the initial Single-Task 3D CNN, the preprocessed lesion-centred CT volume of size $1 \times 64 \times 64 \times 64$ was used as the input to the baseline model. Since pulmonary nodules extend across multiple CT slices, a 3D convolutional neural network was used so that the model could learn volumetric information across depth, height, and width rather than relying on a single 2D cross-section \cite{kang2017_3dmvcnn}.

A compact 3D CNN was used as the baseline model. The network consisted of three 3D convolutional layers with 16, 32, and 64 output channels, respectively. Each convolutional layer used a $3 \times 3 \times 3$ kernel with stride 1 and padding 1, followed by a rectified linear unit (ReLU) activation. Max-pooling with a $2 \times 2 \times 2$ kernel and stride 2 was applied after the first and second convolutional layers to progressively reduce the spatial dimensions.

After the final convolutional layer, adaptive average pooling was applied to reduce each of the 64 feature maps to a single value, resulting in a 64-dimensional feature representation. This representation was passed to a linear layer with one output, producing a single raw logit for binary malignancy-risk prediction. The model did not use dropout or normalization layers.

The model was trained using binary cross-entropy with logits loss. A positive-class weight based on the ratio of negative to positive training samples was used to account for class imbalance. During evaluation, a sigmoid transformation was applied to the output logit to obtain the predicted probability of the higher-risk class. The single-task model contained 69,729 trainable parameters. 


\subsection{Multi-Task 3D CNN}

The encoder used in the single-task 3D CNN was also used in the multi-task 3D CNN without changing its architecture. Each lesion-centred CT volume of size $1 \times 64 \times 64 \times 64$ was passed through three convolutional layers with 16, 32, and 64 output channels, respectively. ReLU activations were applied after each convolutional layer, with max-pooling after the first two layers. Adaptive average pooling was then applied to obtain a shared 64-dimensional feature representation.

\begin{figure*}[!t]
\centering

\begin{tikzpicture}[
    font=\footnotesize,
    block/.style={
        rectangle,
        draw,
        rounded corners,
        minimum width=2.7cm,
        minimum height=0.8cm,
        align=center,
        inner sep=4pt
    },
    output/.style={
        rectangle,
        draw,
        rounded corners,
        minimum width=2.4cm,
        minimum height=0.75cm,
        align=center,
        inner sep=4pt
    },
    arrow/.style={
        -{Latex[length=1.8mm]},
        thick
    }
]


\node[block] (input)
{Lesion-centred CT crop\\
$1 \times 64 \times 64 \times 64$};

\node[block, right=1.0cm of input] (encoder)
{Common 3D Encoder};

\node[block, right=1.0cm of encoder] (features)
{64-dimensional\\
feature representation};

\draw[arrow] (input) -- (encoder);
\draw[arrow] (encoder) -- (features);


\node[output,
      right=1.7cm of features,
      yshift=1.25cm]
      (singlehead)
{Linear $64 \rightarrow 1$};

\node[output,
      right=0.7cm of singlehead]
      (singleout)
{Malignancy-risk\\logit};

\node[
      above=0.35cm of singlehead,
      font=\footnotesize\bfseries]
      (singlelabel)
{Single-Task Model};

\draw[arrow]
(features.east) -- ++(0.55,0)
|- (singlehead.west);

\draw[arrow]
(singlehead) -- (singleout);


\node[output,
      right=1.7cm of features,
      yshift=-0.65cm]
      (malhead)
{Linear $64 \rightarrow 1$};

\node[output,
      below=0.45cm of malhead]
      (spichead)
{Linear $64 \rightarrow 1$};

\node[output,
      below=0.45cm of spichead]
      (lobhead)
{Linear $64 \rightarrow 1$};

\node[output,
      right=0.7cm of malhead]
      (malout)
{Malignancy-risk\\logit};

\node[output,
      right=0.7cm of spichead]
      (spicout)
{Spiculation\\logit};

\node[output,
      right=0.7cm of lobhead]
      (lobout)
{Lobulation\\logit};

\node[
      above=0.35cm of malhead,
      font=\footnotesize\bfseries]
      (multilabel)
{Multi-Task Model};

\draw[arrow]
(features.east) -- ++(0.55,0)
|- (malhead.west);

\draw[arrow]
(features.east) -- ++(0.55,0)
|- (spichead.west);

\draw[arrow]
(features.east) -- ++(0.55,0)
|- (lobhead.west);

\draw[arrow] (malhead) -- (malout);
\draw[arrow] (spichead) -- (spicout);
\draw[arrow] (lobhead) -- (lobout);

\end{tikzpicture}

\caption{Overview of the single-task and multi-task 3D CNN models.
Both models use the same 3D convolutional encoder, summarized in
Table~\ref{tab:common_encoder}, to obtain a 64-dimensional feature
representation from each lesion-centred CT volume. The single-task model
uses one prediction head for malignancy risk, whereas the multi-task model
uses separate heads for malignancy risk, spiculation, and lobulation.}

\label{fig:model_architectures}
\end{figure*}
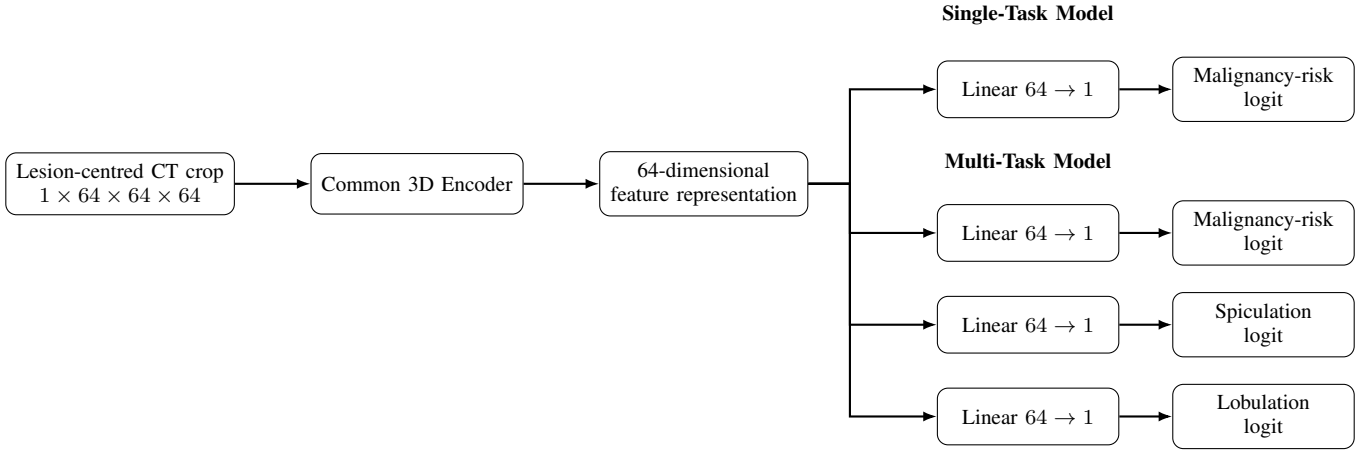

The final layer used three independent linear prediction heads, with one head for malignancy risk and two auxiliary heads for the morphological features spiculation and lobulation. Each head produced a single raw logit corresponding to a binary reader-assessed target. The same rating scheme was applied across all three tasks, with ratings of 1 and 2 assigned to the lower class and ratings of 4 and 5 assigned to the higher class. An overview of the two model
architectures is shown in Fig.~\ref{fig:model_architectures}.

The common encoder used by both models is summarized in
Table~\ref{tab:common_encoder}.

\begin{table}[htbp]
\caption{Common 3D Encoder Architecture}
\centering
\small
\begin{tabularx}{\columnwidth}{l X l}
\toprule
\textbf{Stage} & \textbf{Operation} & \textbf{Output} \\
\midrule

Input
& Lesion-centred CT crop
& $1 \times 64 \times 64 \times 64$ \\

Conv 1
& $3 \times 3 \times 3$, 16 channels + ReLU
& $16 \times 64 \times 64 \times 64$ \\

Pool 1
& $2 \times 2 \times 2$ max pooling
& $16 \times 32 \times 32 \times 32$ \\

Conv 2
& $3 \times 3 \times 3$, 32 channels + ReLU
& $32 \times 32 \times 32 \times 32$ \\

Pool 2
& $2 \times 2 \times 2$ max pooling
& $32 \times 16 \times 16 \times 16$ \\

Conv 3
& $3 \times 3 \times 3$, 64 channels + ReLU
& $64 \times 16 \times 16 \times 16$ \\

Avg. Pool
& Adaptive global average pooling
& $64$ \\

\bottomrule
\end{tabularx}
\label{tab:common_encoder}
\end{table}

For the auxiliary tasks, reader ratings of 3 or unavailable labels were excluded from the respective task loss, while the same samples were retained for malignancy-risk prediction. Each task used binary cross-entropy with logits loss with a task-specific positive-class weight calculated from the ratio of negative to positive training samples among the valid labels available for that task.

The overall training loss was calculated as the mean of the task losses available in each batch. Malignancy-risk loss was retained for all included samples, whereas an auxiliary task did not contribute to the total loss when no valid labels were available for that task within a batch. Sigmoid was applied only during prediction to convert each output logit into a task-specific probability. The multi-task model contained 69,859 trainable parameters.


\subsection{Training and Validation Strategy}

Both models were implemented using PyTorch \cite{paszke2019pytorch}. As mentioned earlier, the single-task and multi-task experiments used the same patient-level split for the training, validation, and test sets. A fixed random seed of 42 was used for dataset partitioning and model training. Python, NumPy, and PyTorch were also seeded with the same value to support reproducibility. A batch size of 2 samples was used during training, with no reshuffling of samples between epochs. 

A learning rate of 0.001 with the Adam optimizer \cite{kingma2015adam} was used to train both models for five epochs. Neither a learning-rate scheduler nor an early-stopping procedure was used. Each training epoch was followed by validation. For the single-task model, the checkpoint with the lowest validation malignancy loss was retained. The multi-task model followed the same selection principle, with checkpoint retention based on the malignancy-risk validation loss rather than the combined loss across all three prediction tasks.

Decision-threshold calculation was performed after checkpoint selection. The retained model was reloaded and predictions were obtained on the validation set. Rather than using a fixed threshold of 0.5, the threshold corresponding to the highest balanced accuracy on the validation set was selected. The same procedure was applied to the multi-task model, but independently for malignancy risk, spiculation, and lobulation using the valid labels available for each task. The held-out test set was then evaluated using the selected checkpoint and the validation-derived threshold values.

\subsection{Evaluation and Statistical Analysis}

Model performance was evaluated at the reader-annotation level using accuracy, balanced accuracy, sensitivity, specificity, positive predictive value (PPV), negative predictive value (NPV), F1 score, receiver operating characteristic area under the curve (ROC-AUC), and average precision. Binary predictions were also analysed using confusion matrices. The held-out test predictions used the decision thresholds derived from the validation set without further tuning. For the multi-task model, the auxiliary tasks were evaluated only for records with valid labels for the corresponding morphological feature.

Uncertainty in the comparison between the two models was assessed using a paired patient-level bootstrap analysis \cite{efron1993bootstrap}. Patients were sampled with replacement, while all associated reader-level annotations for each sampled patient were retained. A total of 2,000 bootstrap resamples were generated using a fixed random seed of 42. For each resample, ROC-AUC, average precision, and balanced accuracy were calculated for both models, together with the paired difference defined as the multi-task metric minus the corresponding single-task metric. The 95\% confidence intervals were estimated using the $2.5^{\mathrm{th}}$ and $97.5^{\mathrm{th}}$ percentiles of the bootstrap distributions.

A subgroup analysis was performed according to CT slice thickness. Reader-level records were divided into two groups based on the DICOM \textit{SliceThickness} value: $\leq 2$ mm and $>2$ mm. The original validation-derived decision thresholds were retained when calculating ROC-AUC and balanced accuracy within each subgroup. Patient counts and reader-level record counts were also recorded for each subgroup. No additional hypothesis testing or p-value calculation was performed.


\section{Results}


\subsection{Cohort Characteristics}

The final cohort consisted of 3,918 reader-level annotations from 742 patients, which were divided into training, validation, and held-out test sets. The training set contained 519 patients and 2,754 annotations, comprising 1,754 lower-risk and 1,000 higher-risk annotations. The validation set contained 111 patients and 536 annotations, including 294 lower-risk and 242 higher-risk annotations. The held-out test set contained the remaining 112 patients and 628 annotations, of which 375 were lower risk and 253 were higher risk. The cohort distribution is summarized in Table~\ref{tab:cohort}.

\begin{table}[htbp]
\caption{Cohort Characteristics and Malignancy-Risk Distribution}
\centering
\small
\begin{tabular}{lrrrr}
\toprule
\textbf{Split} & \textbf{Patients} & \textbf{Annotations} & \textbf{Negative} & \textbf{Positive} \\
\midrule
Training   & 519 & 2754 & 1754 & 1000 \\
Validation & 111 & 536  & 294  & 242  \\
Test       & 112 & 628  & 375  & 253  \\
\midrule
Total      & 742 & 3918 & 2423 & 1495 \\
\bottomrule
\end{tabular}
\label{tab:cohort}
\end{table}

\subsection{Single-Task Malignancy Classification}

For the single-task model, the validation-derived decision threshold for malignancy-risk classification was 0.492. On the held-out test set, the model achieved a ROC-AUC of 0.552 and an average precision of 0.431. The balanced accuracy was 0.548, while the overall accuracy was 0.490. Sensitivity was relatively high at 0.842, whereas specificity was 0.253. The corresponding PPV, NPV, and F1 score were 0.432, 0.704, and 0.571, respectively. At the selected decision threshold, the model produced 95 true negatives, 280 false positives, 40 false negatives, and 213 true positives.


\subsection{Multi-Task Malignancy Classification}

For the malignancy-risk head of the multi-task model, the validation-derived decision threshold was 0.503. On the held-out test set, the model achieved a ROC-AUC of 0.558 and an average precision of 0.443. The balanced accuracy was 0.539, with an overall accuracy of 0.484. Sensitivity was 0.822 and specificity was 0.256. The corresponding PPV, NPV, and F1 score were 0.427, 0.681, and 0.562, respectively. At the selected decision threshold, the model produced 96 true negatives, 279 false positives, 45 false negatives, and 208 true positives.

\subsection{Auxiliary Morphological Concept Prediction}

For the auxiliary morphological tasks, 584 test records from the cohort had valid spiculation labels and 557 had valid lobulation labels. The validation-derived thresholds were 0.503 and 0.523 for spiculation and lobulation, respectively. Spiculation prediction achieved a ROC-AUC of 0.546 and an average precision of 0.137, with a balanced accuracy of 0.499. Sensitivity was 0.030 and specificity was 0.969, while the F1 score was 0.047. For lobulation, the ROC-AUC was 0.480 and the average precision was 0.088, with a balanced accuracy of 0.464. Sensitivity and specificity were 0.560 and 0.369, respectively, and the F1 score was 0.141. Detailed performance metrics for both auxiliary tasks are summarized in Table~\ref{tab:auxiliary_performance}.

\begin{table}[!t]
\caption{Auxiliary Morphological Concept Prediction Performance}
\centering
\small
\begin{tabular}{lcc}
\toprule
\textbf{Metric} & \textbf{Spiculation} & \textbf{Lobulation} \\
\midrule
Valid records & 584 & 557 \\
Positive records & 67 & 50 \\
Threshold & 0.503 & 0.523 \\
ROC-AUC & 0.546 & 0.480 \\
Average Precision & 0.137 & 0.088 \\
Balanced Accuracy & 0.499 & 0.464 \\
Sensitivity & 0.030 & 0.560 \\
Specificity & 0.969 & 0.369 \\
F1 Score & 0.047 & 0.141 \\
\bottomrule
\end{tabular}
\label{tab:auxiliary_performance}
\end{table}


\subsection{Single-Task vs. Multi-Task Comparison}

\begin{figure}[t]
\centering

\includegraphics[width=\columnwidth]{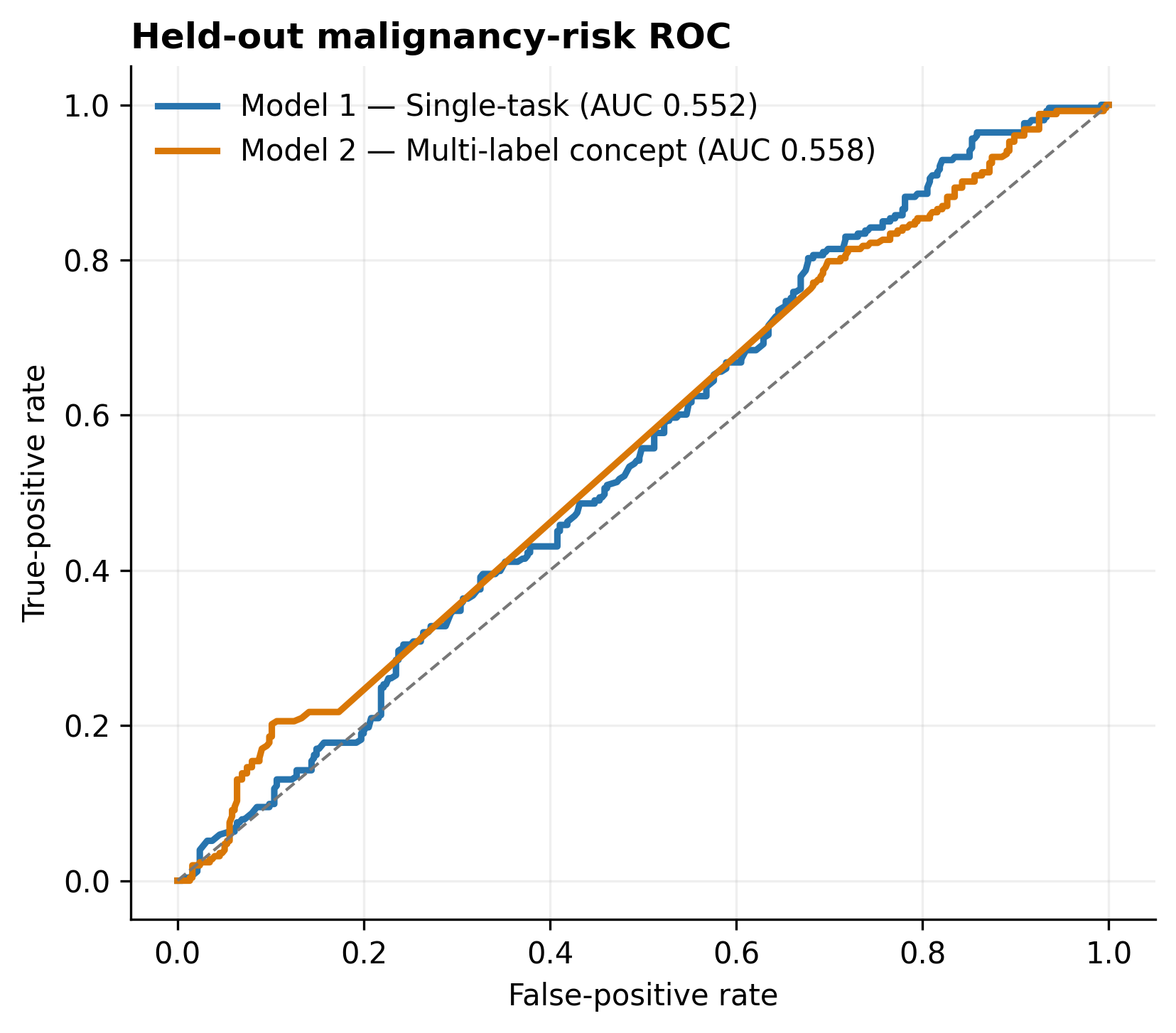}

\vspace{0.6em}

\includegraphics[width=\columnwidth]{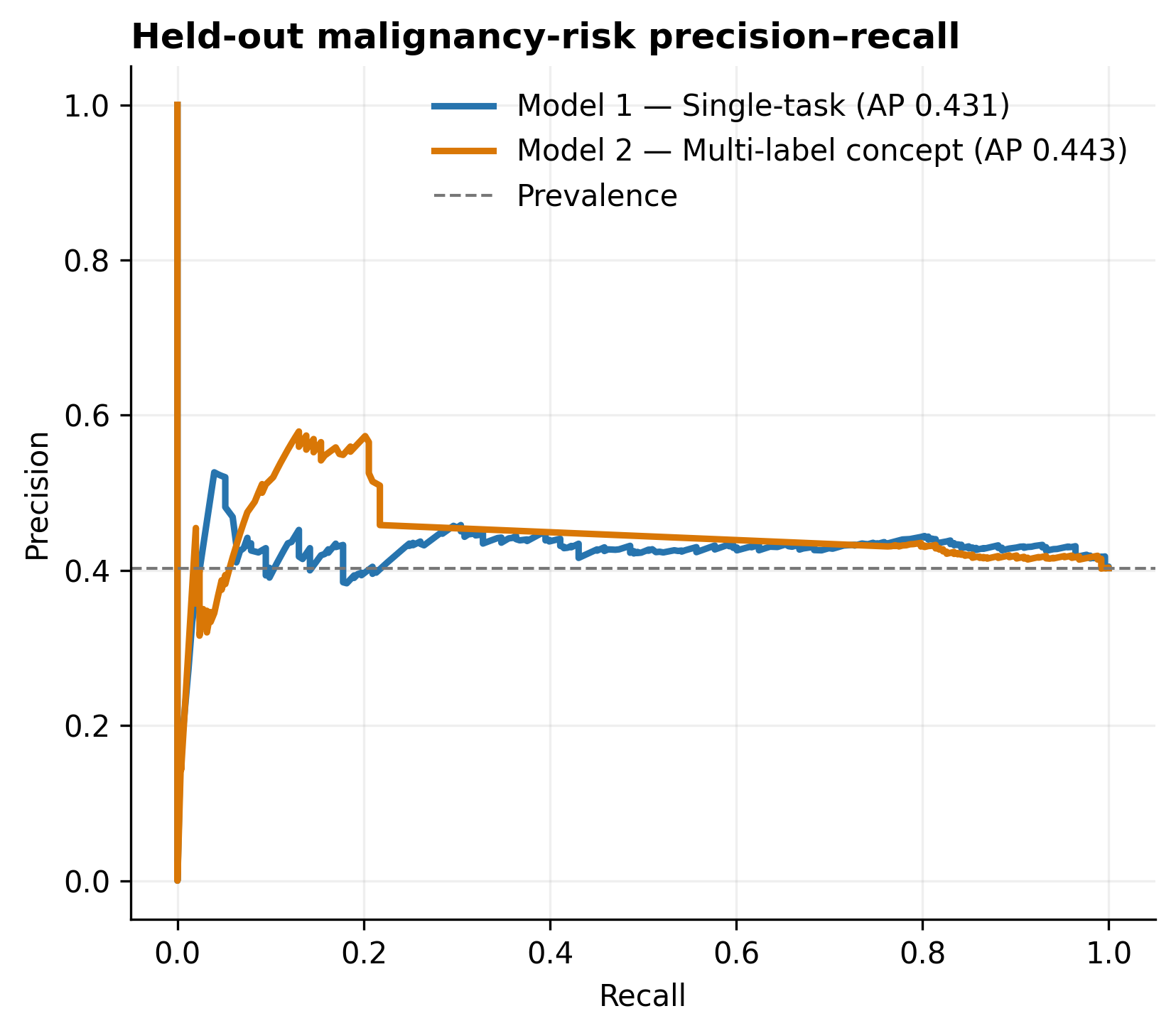}

\caption{Comparison of single-task and multi-task malignancy-risk classification on the held-out test set. The upper panel shows the receiver operating characteristic (ROC) curves, and the lower panel shows the precision--recall curves. AP denotes average precision.}
\label{fig:model_comparison_curves}
\end{figure}

The multi-task model showed a small increase in ROC-AUC and average precision in comparison with the single-task model, with differences of +0.005 and +0.012, respectively. In contrast, balanced accuracy was slightly lower for the multi-task model, decreasing from 0.548 to 0.539. Sensitivity also decreased from 0.842 to 0.822, while specificity remained similar between the two models at 0.253 and 0.256. The remaining performance differences were small, as summarized in Table~\ref{tab:model_comparison}. The ROC and precision--recall curves for the two malignancy-risk models are shown in Fig.~\ref{fig:model_comparison_curves}.

\begin{table}[!t]
\caption{Comparison of Malignancy-Risk Classification Performance}
\centering
\small
\begin{tabular}{lccc}
\toprule
\textbf{Metric} & \textbf{Single-Task} & \textbf{Multi-Task} & \textbf{Difference} \\
\midrule
ROC-AUC & 0.552 & 0.558 & +0.005 \\
Average Precision & 0.431 & 0.443 & +0.012 \\
Balanced Accuracy & 0.548 & 0.539 & -0.009 \\
Accuracy & 0.490 & 0.484 & -0.006 \\
Sensitivity & 0.842 & 0.822 & -0.020 \\
Specificity & 0.253 & 0.256 & +0.003 \\
PPV & 0.432 & 0.427 & -0.005 \\
NPV & 0.704 & 0.681 & -0.023 \\
F1 Score & 0.571 & 0.562 & -0.009 \\
\bottomrule
\end{tabular}
\label{tab:model_comparison}
\end{table}

\subsection{Patient-Level Bootstrap Analysis}

A total of 2,000 valid resamples were produced by the paired patient-level bootstrap analysis. The difference in ROC-AUC between the multi-task and single-task models was +0.005, with a 95\% confidence interval of $-0.087$ to $0.090$. For average precision, the corresponding difference was +0.012, with a 95\% confidence interval of $-0.071$ to $0.093$. The difference in balanced accuracy was $-0.009$, with a 95\% confidence interval of $-0.067$ to $0.043$. All three confidence intervals included zero.

\subsection{Slice-Thickness Subgroup Analysis}

The held-out test set was also examined according to CT slice thickness. There were 68 patients and 366 reader-level records in the $\leq 2$ mm group, compared with 44 patients and 262 records in the $>2$ mm group. For scans with a slice thickness of $\leq 2$ mm, the single-task model obtained a ROC-AUC of 0.555 and a balanced accuracy of 0.555, while the multi-task model obtained 0.536 and 0.539, respectively. In the $>2$ mm group, the ROC-AUC was 0.539 for the single-task model and 0.586 for the multi-task model, while the corresponding balanced accuracies were 0.531 and 0.519. The validation-derived decision thresholds were kept unchanged for both subgroups. The subgroup results are summarized in Table~\ref{tab:slice_thickness}, where AUC denotes ROC-AUC and BA denotes balanced accuracy.

\begin{table}[htbp]
\caption{Performance by CT Slice-Thickness Subgroup}
\centering
\small
\begin{tabular}{lcccc}
\toprule
\textbf{Slice} & \textbf{Patients} & \textbf{Records} &
\textbf{Single-Task} & \textbf{Multi-Task} \\
 &  &  & \textbf{AUC / BA} & \textbf{AUC / BA} \\
\midrule
$\leq 2$ mm & 68 & 366 & 0.555 / 0.555 & 0.536 / 0.539 \\
$>2$ mm     & 44 & 262 & 0.539 / 0.531 & 0.586 / 0.519 \\
\bottomrule
\end{tabular}
\label{tab:slice_thickness}
\end{table}


\section{Discussion}
The main finding of this study was that adding morphological features such as spiculation and lobulation did not produce a clear improvement in malignancy-risk classification compared with the single-task model. Although the multi-task model incorporated two additional morphological prediction tasks, the differences between the two models remained small across the main performance metrics. The multi-task model showed slightly higher ROC-AUC and average precision, whereas balanced accuracy, sensitivity, and F1 score were slightly lower. Together with the patient-level bootstrap results, this suggests that the additional morphological information did not provide a consistent benefit to malignancy-risk prediction in the present experimental setting.

One possible reason for this is the difficulty of learning the morphological characteristics themselves. The auxiliary labels were strongly imbalanced, with relatively few positive examples for spiculation and lobulation, and the corresponding prediction performance was limited. If these characteristics are not learned reliably, the shared representation may receive only a weak additional signal from the auxiliary tasks. This does not necessarily mean that morphological information is unhelpful. Rather, more reliable morphological annotations, improved handling of class imbalance, or a formulation that captures these characteristics more effectively may allow them to contribute more meaningfully to malignancy-risk assessment.

The findings of the present study differ from previous work in which
morphological or semantic information was incorporated into malignancy
prediction frameworks. Shen et al. reported benefits from combining
radiologist-interpreted semantic characteristics with malignancy prediction,
while Wang et al. incorporated morphological and density information as
auxiliary tasks within a multi-task learning framework
\cite{shen2019semantic,wang2022deepln}. In the present study, however,
including spiculation and lobulation as auxiliary prediction tasks did not
produce a clear improvement in the primary malignancy-risk classification.
The differences observed across the main performance metrics were small and
did not indicate a consistent advantage for the multi-task model. These
different outcomes may be influenced by factors such as model architecture,
auxiliary-label formulation, class imbalance, and the way semantic or
morphological information is incorporated into the learning framework.

An important motivation for including morphological characteristics is that they can provide context alongside the predicted malignancy risk. In a clinical setting, a model is more useful when its output can support, rather than replace, the clinician's assessment. A malignancy-risk prediction accompanied by information about relevant imaging characteristics could potentially help a clinician understand which visual features are associated with the model's assessment and may make the output easier to interpret in the context of the CT image. From this perspective, morphological concept learning remains worth investigating even though it did not improve the primary classification task in the present experiment.

Another direction worth exploring is whether morphological concept learning can help when models are evaluated across more heterogeneous or demographically different datasets. If features such as spiculation and lobulation can be predicted reliably across cohorts, they may provide additional context for understanding whether the relationship between imaging appearance and reader-assessed malignancy risk changes across populations or imaging conditions. This could help identify which morphological characteristics remain consistently associated with higher malignancy-risk annotations and which relationships vary across datasets.

The slice-thickness analysis also indicated some variation in model performance across acquisition groups. In the $>2$ mm subgroup, the multi-task model showed a higher ROC-AUC than the single-task model, whereas this pattern was not observed for the $\leq 2$ mm subgroup. However, the subgroup sizes were smaller than the complete test cohort, and the analysis was exploratory. These findings therefore do not establish an effect of slice thickness, but they support further investigation of acquisition heterogeneity when evaluating model generalization.


\section{Limitations}
This study has several limitations. First, the prediction target is based on reader-assessed malignancy ratings rather than histopathologically confirmed diagnosis. Because different radiologists may assign different scores to the same lesion, the annotations themselves can contain reader-level variability and uncertainty. Patient-level splitting was used to prevent overlap of the same patient across the training, validation, and test sets, but multiple reader annotations may still correspond to the same underlying lesion. The model should therefore be interpreted as predicting radiologist-assessed malignancy risk rather than confirming whether a patient has cancer.

A second limitation is the class definition used in the binary experiment. Although the original annotations contain malignancy ratings from 1 to 5, rating 3 was excluded to separate lower-risk and higher-risk cases more clearly. This simplifies the classification problem, but it also removes indeterminate cases that may be clinically important and could influence the observed performance if handled differently.

The auxiliary morphological tasks were also limited by the available labels. Positive examples for spiculation and lobulation were relatively uncommon, and some annotations were unavailable or excluded from the corresponding auxiliary loss. This makes it difficult to determine whether the limited improvement from multi-task learning is mainly due to the formulation itself or to the uncertainty and imbalance present in the auxiliary targets.

Finally, all experiments were performed using a single public dataset, and the model was not externally validated on data from a different institution or acquisition setting. Differences in scanner characteristics, reconstruction protocols, population characteristics, or annotation practices may affect model behaviour. The current results should therefore not be assumed to generalize directly to other datasets or clinical settings without further validation.


\section{Conclusion}

This study examined whether learning radiologist-defined morphological characteristics such as spiculation and lobulation alongside malignancy risk could improve classification from 3D CT images compared with a single-task model. In the current setting, the multi-task approach did not provide a clear and consistent improvement in malignancy-risk prediction, and the auxiliary tasks were themselves limited by class imbalance and limitations in the available auxiliary annotations. This suggests that simply adding morphological prediction heads may not be enough unless those features are learned from stronger and more reliable supervision. Future work could benefit from better handling of auxiliary labels, external validation across more heterogeneous datasets, and the use of histopathologically confirmed outcomes where available, alongside radiologist annotations, to provide a stronger reference for understanding how morphological characteristics relate to malignancy. Such an approach could help determine whether interpretable imaging features can contribute more reliably to clinically useful malignancy-risk assessment.

\section*{Acknowledgment}

The author acknowledges the National Cancer Institute and the Foundation for the National Institutes of Health, and their critical role in the creation of the free publicly available LIDC/IDRI Database used in this study.


\end{document}